\documentclass[11pt]{article}
\usepackage[margin=1in]{geometry}
\usepackage{amsmath,amssymb}
\usepackage{booktabs}
\usepackage{array}
\usepackage{tabularx}
\usepackage{float}
\newfloat{algorithm}{tbp}{loa}
\floatname{algorithm}{Algorithm}
\usepackage{graphicx}
\usepackage{xcolor}
\usepackage{indentfirst}
\usepackage{hyperref}
\usepackage[authoryear,round]{natbib}

\hypersetup{
  colorlinks=true,
  linkcolor=blue,
  citecolor=blue,
  urlcolor=blue,
  pdftitle={Ontology-Grounded World Models for Failure Diagnosis and Closed-Loop Repair in Physical AI Systems},
  pdfauthor={Kailin Wang, Haoxiang Jie, Yaoyuan Yan, Jiacheng Zhou, Zhiyou Heng},
  pdfkeywords={Operational Robot Task Ontology, Physical AI Systems, World Models, Typed Failure Diagnosis, Verification-Gated Correction}
}
\renewcommand{\arraystretch}{1.16}
\newcolumntype{Y}{>{\raggedright\arraybackslash}X}
\newcolumntype{L}[1]{>{\raggedright\arraybackslash}p{#1}}
\newcolumntype{C}[1]{>{\centering\arraybackslash}p{#1}}
\newcommand{\best}[1]{\textbf{#1}}

\makeatletter
\def\ps@preprint{%
  \def\@oddhead{}%
  \def\@evenhead{}%
  \def\@oddfoot{\hbox to\textwidth{\hfil\thepage\hfil}}%
  \def\@evenfoot{\hbox to\textwidth{\hfil\thepage\hfil}}%
}
\let\ps@plain\ps@preprint
\makeatother
\makeatletter
\renewcommand{\@maketitle}{%
  \newpage
  \null
  \vspace*{-0.35in}%
  \begin{center}%
    \fontfamily{ptm}\selectfont
    \rule{0.88\textwidth}{1.1pt}\par
    \vspace{1.1em}%
    {\LARGE\bfseries\centering
      \begin{minipage}{0.92\textwidth}
      \centering\@title
      \end{minipage}\par}%
    \vspace{0.95em}%
    \rule{0.88\textwidth}{1.1pt}\par
    \vspace{1.15em}%
    {\normalsize \@author\par}%
  \end{center}%
  \vspace{1.0em}%
}
\makeatother

\title{Ontology-Grounded World Models for Failure Diagnosis and Closed-Loop Repair in Physical AI Systems}
\author{
\textbf{Kailin Wang\textsuperscript{1}} \quad
\textbf{Haoxiang Jie\textsuperscript{1}} \quad
\textbf{Yaoyuan Yan\textsuperscript{1}} \quad
\textbf{Jiacheng Zhou\textsuperscript{2,3}} \quad
\textbf{Zhiyou Heng\textsuperscript{1}}\\[0.35em]
{\small \textsuperscript{1}AI Lab, Country Garden Services Group \quad
\textsuperscript{2}Fudan University \quad
\textsuperscript{3}Omni AI}
}
\date{}

\begin{document}
\newcommand{\LiberoSeedCount}{4}
\newcommand{\LiberoWindowsPerSeed}{500}
\newcommand{\LiberoTotalWindows}{2000}
\newcommand{\LiberoPooledBeforeCount}{1739}
\newcommand{\LiberoPooledAfterCount}{1881}
\newcommand{\LiberoBaselineFailureCount}{261}
\newcommand{\LiberoRescuedCount}{142}
\newcommand{\LiberoBeforeMeanPct}{86.95}
\newcommand{\LiberoBeforeStdPct}{1.00}
\newcommand{\LiberoAfterMeanPct}{94.05}
\newcommand{\LiberoAfterStdPct}{0.30}
\newcommand{\LiberoRecoveryPct}{54.4}
\newcommand{\LiberoGainMeanPct}{7.10}
\newcommand{\LiberoGainStdPct}{1.01}
\newcommand{\LiberoSeedZeroBefore}{436}
\newcommand{\LiberoSeedZeroAfter}{469}
\newcommand{\LiberoSeedZeroFailures}{64}
\newcommand{\LiberoSeedZeroRescued}{33}
\newcommand{\LiberoSeedOneBefore}{428}
\newcommand{\LiberoSeedOneAfter}{471}
\newcommand{\LiberoSeedOneFailures}{72}
\newcommand{\LiberoSeedOneRescued}{43}
\newcommand{\LiberoSeedTwoBefore}{435}
\newcommand{\LiberoSeedTwoAfter}{469}
\newcommand{\LiberoSeedTwoFailures}{65}
\newcommand{\LiberoSeedTwoRescued}{34}
\newcommand{\LiberoSeedThreeBefore}{440}
\newcommand{\LiberoSeedThreeAfter}{472}
\newcommand{\LiberoSeedThreeFailures}{60}
\newcommand{\LiberoSeedThreeRescued}{32}

\maketitle

\begin{abstract}
EV-WM represents candidate quality with feature and event scores, but these scores do not explicitly record an unmet task predicate, a route label for an available correction mechanism, or a post-correction acceptance result. We present \textbf{Onto-EV-WM}, an ontology-grounded diagnosis and verification-gated correction interface layered above EV-WM rather than a replacement world-model architecture. The implemented task-local TBox defines entity types, predicate signatures, and constraints; source-specific grounding maps predicted or simulator-observed states to task ABoxes; and deterministic rules retain each missing predicate and its arguments when assigning a route label. Learned or heuristic proposers remain separate from this symbolic interface; native task predicates determine acceptance, and the bounded protocol determines whether a failed verification is retried. In the aligned PointMaze evaluation, EV-WM and Onto-EV-WM both report 94\% success, with mean final-state distances of 0.90573 and 0.61177, respectively; the separately budgeted search reaches 100\% success. On LIBERO-Goal, the ontology represents failed task conditions as typed records, retains their predicate arguments, and associates them with the declared source/joint correction route and predicate-gated acceptance; the complete configuration reports 93.8\% corrected-window success on seed 0 and $\LiberoAfterMeanPct\!\pm\!\LiberoAfterStdPct\%$ across four evaluation-sampling seeds. On the fixed 10,030-task LIBERO-Plus registry, Onto-EV-WM succeeds on 8,526 tasks (85.00\%), with suite-level success rates of 65.98\% for LIBERO-10, 91.39\% for LIBERO-Goal, and 91.38\% for both LIBERO-Object and LIBERO-Spatial. These numbers report the performance of the complete ontology-grounded configurations under the tested simulator protocols; an ontology-only causal share is not measured separately, and real-robot recovery is not evaluated.
\end{abstract}

\section{Introduction}

We consider embodied systems that predict candidate futures, apply actions through a controller, and evaluate task success under physical and task constraints. In the systems studied here, a world model supplies and scores candidate rollouts \citep{ha2018worldmodels,hafner2019planet,hafner2020dreamer,hafner2023dreamerv3,oquab2023dinov2,nvidia2025cosmos}, while separate components perform predicate diagnosis and post-application verification.

In predicate-based manipulation benchmarks, visual or feature-space proximity is not equivalent to task success. For example, an imagined state may place a plate near the stove without satisfying \texttt{in\_region(plate, stove\_front)}. A scalar candidate score does not specify which required predicate is unsatisfied; the unmet condition may instead be a spatial relation, joint state, or contact constraint. In our formulation, the correction interface therefore records the failed condition, assigns a declared route label, and queries the task predicate after application.

The Event-Verified World Models (EV-WM) framework decodes imagined futures into task-event scores and uses them to rank candidate windows \citep{wang2026evwm}. Its event vector does not explicitly retain a failed predicate with typed arguments or define post-correction retry. Onto-EV-WM adds a task-local record for grounding, diagnosis, route assignment, application mode, and verification.

Our aim is not to replace the underlying world model or visuomotor controller with a new policy class. Instead, ontology grounding provides an explicit interface between prediction, correction mechanisms, and verification. It records the task predicate, typed arguments, threshold margin, diagnosed predicate failure, selected correction family, application mode, and verification outcome. Throughout the paper, a \emph{route} is a dispatch label for a compatible mechanism rather than a robot action. Direct simulator-state application and controller-mediated execution are therefore described separately. We use the term \emph{Physical AI system} in this system-level sense; all experiments are simulation-based and do not constitute real-robot or sim-to-real validation.

We introduce \textbf{Onto-EV-WM}, which maps predicted outputs or simulator states to typed, task-specific assertions in an operational robot task ontology. Its TBox defines the entity and relation vocabulary, predicate signatures, and type constraints. A separate deterministic rule base diagnoses missing assertions and maps typed failures to task-specific correction routes. At run time, the task specification and a predicted or simulator-observed state instantiate source-specific ABoxes of required and observed facts. A missing required assertion becomes a typed failure instance without discarding the original predicate arguments. In LIBERO-Goal, the quantitative protocol uses one fixed source/joint \texttt{qpos}-delta mechanism after replay failure. In this paper, the ontology is implemented as a task-local schema and deterministic rule set, not as a large external knowledge base, an open-world knowledge graph, or a replacement world model.

This paper makes three contributions:
\begin{itemize}
\item We design an operational robot task ontology and an explicit grounding interface that represent predicted and simulator-observed states as distinct, task-specific ABoxes under a shared vocabulary and type constraints.
\item We define deterministic diagnosis rules and task-specific route mappings that preserve a violated predicate, its typed arguments, and a threshold margin when defined, and assign route labels associated with the mechanisms available in each protocol.
\item We define a verification-gated correction contract that records diagnosis, route selection, proposal generation, application mode, and native-predicate acceptance or retry.
\end{itemize}

We report three benchmark settings with distinct units: an aligned 50-trial PointMaze comparison, sampled-window LIBERO-Goal correction, and the fixed LIBERO-Plus registry. The four LIBERO-Goal seeds vary evaluation sampling; they are not independent training runs, a held-out split, or full-episode policy evaluation.

\section{Related Work}

\subsection{World Models for Physical AI}

World-model systems learn compact dynamics for control from high-dimensional observations \citep{ha2018worldmodels}. PlaNet plans with latent dynamics \citep{hafner2019planet}, while Dreamer-style agents learn through latent imagination \citep{hafner2020dreamer,hafner2021dreamerv2,hafner2023dreamerv3}; DayDreamer extends this direction toward physical robots \citep{wu2022daydreamer}. World foundation models have also been positioned as predictive infrastructure for Physical AI systems \citep{nvidia2025cosmos}. Feature- and video-prediction methods model future images, video, or learned representations rather than relying exclusively on pixel-level reconstruction \citep{assran2023ijepa,bardes2024vjepa,bruce2024genie,wu2024ivideogpt}. EV-WM is the direct predecessor: it predicts candidate futures in visual-feature space, decodes event states, and evaluates progress, consistency, feasibility, and uncertainty \citep{wang2026evwm}. Onto-EV-WM retains that substrate and adds ontology-grounded diagnosis and correction routing.

These systems differ in how predicted futures become control decisions. Latent-control agents typically optimize value or imagined return, while feature-prediction planners can select futures by representation distance. EV-WM augments this selection process with task-event scores. In the aligned PointMaze configuration, the predictive backbone remains fixed while the typed-fact head and task-local interface are added; the analysis focuses on the record produced when a candidate violates a task condition.

\subsection{Predicate Grounding and Robot Knowledge}

Manipulation success is often defined by symbolic task predicates: an object must be on a target, inside a container, or aligned with a region, and an articulated joint must reach a target state. SayCan grounds instructions in affordance scores \citep{ahn2022saycan}, while RT-1, Diffusion Policy, OpenVLA, and $\pi_0$ generate robot actions from learned visuomotor models \citep{brohan2022rt1,chi2023diffusionpolicy,kim2024openvla,black2024pi0}. Onto-EV-WM does not replace these policies; the present experiments evaluate predicate diagnosis and simulator-state correction.

Direct policies and diagnostic interfaces answer different questions. A policy must produce an action sequence under environment dynamics. The diagnostic interface studied here records why a candidate state does not satisfy the task. Its state-correction results should not be interpreted as action execution.

Although \emph{ontology} originates as a philosophical account of entities and their categories, knowledge engineering gave the term an operational computational meaning. Gruber defined an ontology as an explicit specification of a conceptualization \citep{gruber1993ontology}. In this computational sense, an ontology supplies not only labels, but also machine-interpretable classes, typed relations, admissible arguments, and explicit constraints.

Robotics adapted these ideas to robots, objects, actions, capabilities, environments, and task constraints. CORA defines a core vocabulary for robotics and automation \citep{prestes2013cora}; KnowRob provides knowledge processing for cognition-enabled robots \citep{tenorth2013knowrob,beetz2018knowrob}; and related surveys and systems describe symbolic structures for task execution and knowledge-enabled perception \citep{paulius2018survey,balint2019robosherlock}. Onto-EV-WM uses a narrower, task-local operational ontology instantiated from task specifications and simulator state. It neither claims a universal robot ontology nor performs general open-world reasoning. In the implemented mapping, the shared vocabulary retains the task predicate and typed arguments from diagnosis through route assignment and verification rather than storing rejection only as an unstructured score.

\subsection{Task and Motion Planning}

Task and motion planning (TAMP) combines discrete task structure with continuous geometric and motion feasibility \citep{garrett2021tamp}. Onto-EV-WM instead grounds an imagined future or simulator state, records a missing task predicate, and assigns a route label defined by the current protocol. Its formal interface permits a TAMP solver, policy, or controller as a route back end, but the experiments instantiate only the mechanisms stated in their protocols and do not evaluate a new TAMP solver.

\subsection{Neuro-Symbolic Robotics Interfaces}

Neuro-symbolic systems combine learned representations with explicit symbolic structure and reasoning \citep{garcez2023neurosymbolic}. Onto-EV-WM adopts a limited form of this separation: neural components predict event or fact scores and may propose corrections, while task-local grounding, diagnosis, routing, and acceptance use explicit predicates and deterministic rules. The resulting ABox records the failed predicate, typed arguments, selected route, and verification outcome between prediction and correction. This record depends on a predefined vocabulary and source-specific grounding. We therefore describe the method as a task-local neuro-symbolic interface, not as a general ontology reasoner or differentiable logic system.

\subsection{Verification and Iterative Correction}

Verifiers and rerankers are used to select among generated candidates. Language-agent systems study iterative feedback, self-refinement, critique, search, and tool use \citep{shinn2023reflexion,madaan2023selfrefine,gou2023critic,yao2023tree,yang2024sweagent}. Onto-EV-WM uses a generate--check--correct organization in which feedback is represented by typed task facts. In the LIBERO correction protocol, post-application acceptance is determined by the native simulator predicate.

A binary success verifier separates accepted and rejected candidates but does not necessarily identify the failed task condition. Onto-EV-WM places typed diagnosis between event prediction and verification. The resulting route label is therefore an interface for organizing corrections rather than evidence that any particular correction can be executed by a robot policy.

\subsection{Robot Failure Recovery and Execution Interfaces}

Robot failure-recovery systems combine monitoring, supervision, replanning, and corrective behavior. RACER uses a language-based supervisor to guide a visuomotor recovery policy \citep{dai2024racer}, while recent VLM--behavior-tree systems combine structured scene representations with pre-execution checks and reactive failure handling \citep{ahmad2025failurehandling}. These systems combine online failure detection with corrective action under their respective evaluation protocols.

Onto-EV-WM defines a common record structure for predicted or simulator-grounded task facts, diagnosis metadata, correction metadata, and verifier outcomes. Quantitative LIBERO-Goal uses a fixed route rather than online route selection; controller-mediated use appears only in the selected qualitative trace.

\section{Problem Formulation}

Let $t$ denote the current time step, $k$ the number of preceding steps retained as history, and $H$ the number of future action steps considered by the model. The slice $x_{i:j}$ denotes the inclusive sequence $(x_i,x_{i+1},\ldots,x_j)$. We consider an observation history $o_{t-k:t}$, a proprioceptive-state history $p_{t-k:t}$, a structured task specification $g$, and a candidate action window $a_{t:t+H-1}$. Thus, $o$ contains camera observations, $p$ contains robot-internal measurements such as joint state or end-effector pose, and $a$ contains $H$ proposed control commands. The present experiments do not encode free-form natural language; $g$ is the task identifier or BDDL-style specification from which required task facts are constructed. A visual encoder $E$ and an action-conditioned world model $F_\theta$ produce
\[
z_t=E(o_t), \qquad
\hat z_{t+H}=F_\theta(z_{t-k:t},p_{t-k:t},a_{t:t+H-1}).
\]
Here, $z_t$ is the encoded visual feature at time $t$; $z_{t-k:t}$ is its history; the hat in $\hat z_{t+H}$ marks a prediction rather than an observation; and $\theta$ denotes the learned parameters of $F$. EV-WM adds the learned event predictor
\[
\hat e_{t+H}=G_\psi(\hat z_{t+H},z_{t-k:t},g,a_{t:t+H-1}),
\]
where $G_\psi$ has the single formal role of predicting the task-event state $\hat e_{t+H}$ at the end of the horizon. A benchmark-specific typed-fact head may additionally produce fact scores,
\[
\hat y^{\mathrm{fact}}_{t+H}
=H_\omega(\hat z_{t+H},z_{t-k:t},g,\hat e_{t+H}),
\]
where $H_\omega$ is distinct from the EV-WM event predictor. Deterministic, task-local grounding then constructs three provenance-specific ABoxes:
\[
\mathcal A_g^\star=\phi_{\mathrm{req}}(g),\qquad
\hat{\mathcal A}_{g,t+H}^{\mathrm{pred}}
=\phi_{\mathcal T}^{\mathrm{pred}}(g,\hat e_{t+H},\hat y^{\mathrm{fact}}_{t+H}),
\]
\[
\mathcal A_{g,t}^{\mathrm{sim}}
=\phi_{\mathcal T}^{\mathrm{sim}}(g,s_t^{\mathrm{sim}}).
\]
When a benchmark does not use $H_\omega$, the typed-score argument of $\phi_{\mathcal T}^{\mathrm{pred}}$ is omitted; simulator grounding always bypasses that learned head.
Here, $\mathcal A_g^\star$ is the required ABox compiled from the structured task specification; $\hat{\mathcal A}^{\mathrm{pred}}$ is grounded from learned predicted outputs; and $\mathcal A^{\mathrm{sim}}$ is grounded directly from simulator state. The implementation looks up enumerated task bindings, checks required record fields, and applies task-specific thresholds; it does not perform general ontology-consistency checking or failure-to-route dispatch. In PointMaze, $H_\omega$ supplies learned typed-fact scores while the predictive backbone remains fixed. In the LIBERO correction analysis, simulator state supplies the observed ABox. These sources use the same declared vocabulary but retain different provenance tags.

Let $\mathcal A^{\mathrm{src}}\in\{\hat{\mathcal A}^{\mathrm{pred}},\mathcal A^{\mathrm{sim}}\}$ retain this provenance tag, and let $\mathrm{Sat}_{\mathcal T}(f,\mathcal A^{\mathrm{src}})$ evaluate a required assertion by its declared polarity, threshold, and source-specific evidence. The unsatisfied set is
\[
\Delta\mathcal F^{\mathrm{src}}
=\{f\in\mathrm{Facts}(\mathcal A_g^\star)
\mid \mathrm{Sat}_{\mathcal T}(f,\mathcal A^{\mathrm{src}})=0\}.
\]
This definition covers an absent assertion, an explicitly false predicate, or a violated continuous margin; ordinary set subtraction would not distinguish those cases. It uses a task-scoped closed-world convention only for predicates declared as required by $g$. Absence of an unrelated fact is not interpreted as false, and the method does not claim general open-world ontology reasoning.

Deterministic diagnosis rules $\Pi_D$ create one record for each unsatisfied assertion,
\[
\mathcal D^{\mathrm{src}}
=\{d_f=(f,c_f,m_f)\mid f\in\Delta\mathcal F^{\mathrm{src}},
\ c_f=\Pi_D(f,\mathcal A^{\mathrm{src}}),
\ m_f=\mathrm{Margin}_{\mathcal T}(f,\mathcal A^{\mathrm{src}})\},
\]
where $c_f$ is the typed failure class and $m_f$ is the threshold margin when the task defines one; otherwise $m_f=\bot$ marks that no continuous margin is defined. In the general interface, a task-specific ordering may select a record $d^\dagger$ when several predicates fail, after which the external mapping $\Pi_R$ returns a declared dispatch label
\[
r=\Pi_R(d^\dagger).
\]
Diagnosis and route mappings are deterministic. A proposal generator $P_r$ may be learned, heuristic, or planning-based, whereas application is either direct simulator-state mutation or controller-mediated execution. Quantitative LIBERO-Goal fixes the source/joint route and one learned head. A \emph{predicate failure} is an unsatisfied assertion in $\mathcal D^{\mathrm{src}}$, a \emph{replay failure} is a negative native predicate before correction, and an \emph{execution failure} is a negative native predicate after application.

For a proposed state correction $q$ that produces state $s'$, where the prime denotes the post-correction state, the protocol records success only when the verifier confirms the corrected state:
\[
\mathrm{AcceptState}(s')=\mathbb I[\mathrm{Verifier}(s')=1].
\]
Here, $\mathbb I[\cdot]$ is an indicator and $\mathrm{Verifier}(s')$ is the task-success check. This indicator evaluates only the resulting simulator state; it does not test whether the proposed transition is dynamically realizable by a robot policy. In LIBERO, $q$ directly changes simulator \texttt{qpos} rather than using \texttt{env.step(action)}; acceptance uses the native \texttt{env.check\_success()} query.

More generally, the verification-gated correction interface maintains an attempt index $b=0,1,\ldots,B-1$, where $B$ is the maximum attempt budget. In this general interface specification, a failed verification may trigger fresh grounding and diagnosis, permitting route selection to be recomputed on a later attempt:
\[
\mathcal D^{(b)}=\mathrm{Diagnose}_{\Pi_D}
(\mathcal A_g^\star,\mathcal A_{g}^{\mathrm{src},(b)}),\qquad
(d^{\dagger,(b)},r^{(b)})=\mathrm{SelectRoute}_{\Pi_R}(\mathcal D^{(b)}),
\]
\[
q^{(b)}=P_{r^{(b)}}(s^{(b)},\mathcal A_g^\star,d^{\dagger,(b)}),\qquad
s^{(b+1)}=\mathrm{Apply}_{r^{(b)}}(s^{(b)},q^{(b)}),
\]
\[
\mathrm{LoopDecision}(s^{(b+1)})=
\begin{cases}
\mathrm{accept}, & \mathrm{Verifier}(s^{(b+1)})=1,\\
\mathrm{retry\ or\ reroute}, & \mathrm{Verifier}(s^{(b+1)})=0.
\end{cases}
\]
The superscript $(b)$ is an iteration label. In this general contract, \emph{retry} requests another proposal under the current route, whereas \emph{reroute} reapplies $\Pi_R$ after regrounding. Neither reported protocol evaluates cross-route switching: quantitative LIBERO-Goal uses one fixed-route proposal per failed window, and Task 76 retries within one controller route. This bounded verify--retry contract is the meaning of ``closed loop'' here, not real-robot feedback-control validation.

For clarity, commas separate function arguments or record fields, braces $\{\cdot\}$ denote a set, square brackets in $\mathbb I[\cdot]$ enclose a truth condition, subscripts identify time, task, route, or semantic role, and a superscript star denotes a required target. These conventions apply to every expression in this section.

\section{Method}

\subsection{Architecture and Relationship to EV-WM}

\begin{figure}[htbp]
\centering
\includegraphics[width=\linewidth]{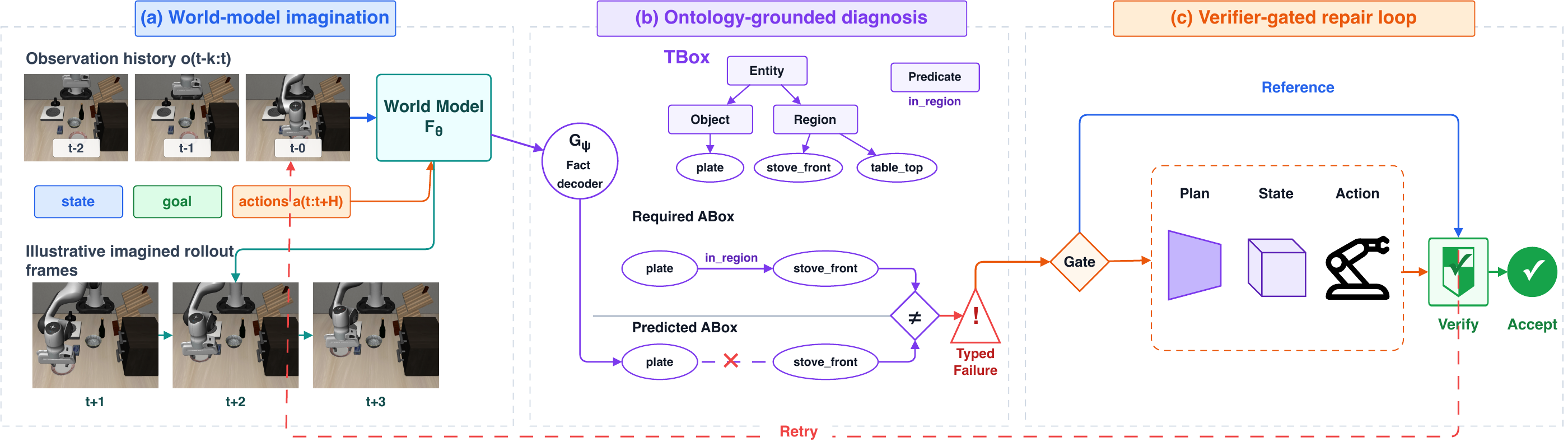}
\caption{Ontology-grounded world-model architecture for failure diagnosis and closed-loop repair. \textbf{a}, Observation, goal, and candidate-action histories condition the world model $F_\theta$, which produces illustrative imagined rollouts. \textbf{b}, The reusable TBox specifies entity types and predicate signatures, while the fact decoder $G_\psi$ instantiates a predicted ABox. Comparing it with the task-required ABox exposes a missing \texttt{in\_region} relation as a typed failure. \textbf{c}, The gate either preserves the reference path or invokes a compatible plan-, state-, or action-level execution mechanism. The verifier accepts a predicate-valid outcome; otherwise, feedback returns to the candidate action window for bounded retry.}
\label{fig:onto_arch}
\end{figure}

Figure~\ref{fig:onto_arch} uses $G_\psi$/``fact decoder'' as a legacy aggregate-stage label, not as the formal ABox-grounding operator. In the formal notation, $G_\psi$ only predicts events; the complete stage additionally contains the optional typed-fact head $H_\omega$ and deterministic grounding $\phi_{\mathcal T}$. Likewise, the diagram's plan-, state-, and action-level labels denote possible external execution back ends; they are not assertions stored in the TBox and do not imply that every back end is quantitatively evaluated here.

The base EV-WM design is retained in Figure~\ref{fig:onto_arch}a: a frozen visual representation supports feature-space rollouts and an event predictor decodes imagined futures into task progress signals \mbox{\citep{wang2026evwm}}. Planning can combine a feature cost with event scores,
\[
J(a_{t:t+H-1})=w_fC_{\mathrm{feature}}-S_{\mathrm{event}}.
\]
For ontology-aware planning, the same candidate set can instead be scored by the task-local verifier,
\[
S_{\mathrm{onto}}=V_{\mathcal T}(\hat{\mathcal A}_{g,t+H}^{\mathrm{pred}},\mathcal A_g^\star),
\qquad
J_{\mathrm{onto}}=w_fC_{\mathrm{feature}}-S_{\mathrm{onto}},
\]
Here, $w_f\geq0$, lower $J_{\mathrm{onto}}$ is preferred, and $V_{\mathcal T}$ denotes the benchmark-configured score computed from the verifier fields enabled in that run. Its active fields and weights are fixed by the benchmark configuration; the paper does not define a benchmark-independent ontology score. Onto-EV-WM adds the ontology-grounded diagnostic and verifier-gated correction interface summarized in Figure~\ref{fig:onto_arch}b--c and Table~\ref{tab:relationship}.

The interface separates prediction, grounding, diagnosis, application, and verification. PointMaze uses predicted facts for candidate scoring; quantitative LIBERO-Goal uses simulator grounding, one fixed correction head, direct \texttt{qpos} application, and native verification; and Task 76 illustrates a within-route controller retry. These are separate instantiations, not one evaluated online system that switches among route types. A satisfied predicted ABox may retain a planning candidate, but only the native predicate records simulator or execution success.

\begin{table}[!htbp]
\centering
\small
\caption{Relationship to EV-WM. ``New'' denotes a component introduced by the ontology-grounded interface, not evidence that the interface alone causes the measured gain.}
\label{tab:relationship}
\begin{tabularx}{0.94\linewidth}{L{0.30\linewidth}cY}
\toprule
Component & Status & Role \\
\midrule
Visual encoder and feature rollout & Reused & Imagine candidate future features \\
Event prediction and scoring & Reused & Estimate task progress and event scores \\
TBox, ABox grounding, and diagnosis & New & Type task assertions and expose missing predicates \\
Failure gate and execution routing & New & Retain the reference or select a plan-, state-, or action-level mechanism \\
Native task predicate & Evaluation tool & Verify the resulting state before acceptance \\
Bounded accept/retry control & New & Terminate on success or update the next candidate window \\
\bottomrule
\end{tabularx}
\end{table}

\begin{algorithm}[H]
\caption{Interface-level pseudocode for source-aware grounding and verification-gated correction}
\label{alg:onto_correction}
\small
\begin{minipage}{0.94\linewidth}
\textbf{Input:} task specification $g$; source tag $\sigma\in\{\mathrm{pred},\mathrm{sim}\}$; predicted output and, when application is requested, state $s^{(0)}$; available route-specific proposers and application modes; native verifier $V$; attempt budget $B$.\\
\textbf{Output:} retained planning candidate, accepted applied state, or unresolved typed failure trace.
\begin{enumerate}
\item Construct the required ABox $\mathcal A_g^\star=\phi_{\mathrm{req}}(g)$ and set $b\leftarrow0$.
\item Ground $\mathcal A_g^{\sigma,(b)}$ with the source-specific $\phi_{\mathcal T}^{\sigma}$, retaining $\sigma$ in the record.
\item Evaluate every required assertion with $\mathrm{Sat}_{\mathcal T}$ and construct $\Delta\mathcal F^{(b)}$. If $\sigma=\mathrm{pred}$ and $\Delta\mathcal F^{(b)}=\varnothing$, retain the reference candidate without reporting executed success. If $\sigma=\mathrm{sim}$, query $V$; return the verified state when $V=1$, and return an unresolved grounding--verifier disagreement when $V=0$ but $\Delta\mathcal F^{(b)}=\varnothing$.
\item Instantiate nonempty $\mathcal D^{(b)}$ from $\Delta\mathcal F^{(b)}$ using deterministic diagnosis rules $\Pi_D$. Select one current mismatch $d^{\dagger,(b)}$ by the declared priority and obtain $r^{(b)}=\Pi_R(d^{\dagger,(b)})$.
\item Generate $q^{(b)}=P_{r^{(b)}}(s^{(b)},\mathcal A_g^\star,d^{\dagger,(b)})$ and apply it using the route's declared simulator-state or controller-mediated contract. After application, set $\sigma\leftarrow\mathrm{sim}$ for observed-state regrounding.
\item Query $V$ on the resulting simulator state. If $V=1$, return the accepted state. Otherwise increment $b$, reground with $\phi_{\mathcal T}^{\mathrm{sim}}$, and retry or reroute from the updated diagnosis while $b<B$.
\item If the budget is exhausted, return the unresolved typed failure trace without asserting task success.
\end{enumerate}
\end{minipage}
\end{algorithm}

\subsection{Ontology-Grounded Diagnosis and Correction Loop}

Algorithm~\ref{alg:onto_correction} is an interface-level specification whose branches need not all be instantiated by a given experiment. Quantitative LIBERO-Goal exercises one fixed-route state proposal per failed replay window; the selected qualitative trace exercises only a within-route controller retry.

\begin{figure}[H]
\centering
\includegraphics[width=\linewidth]{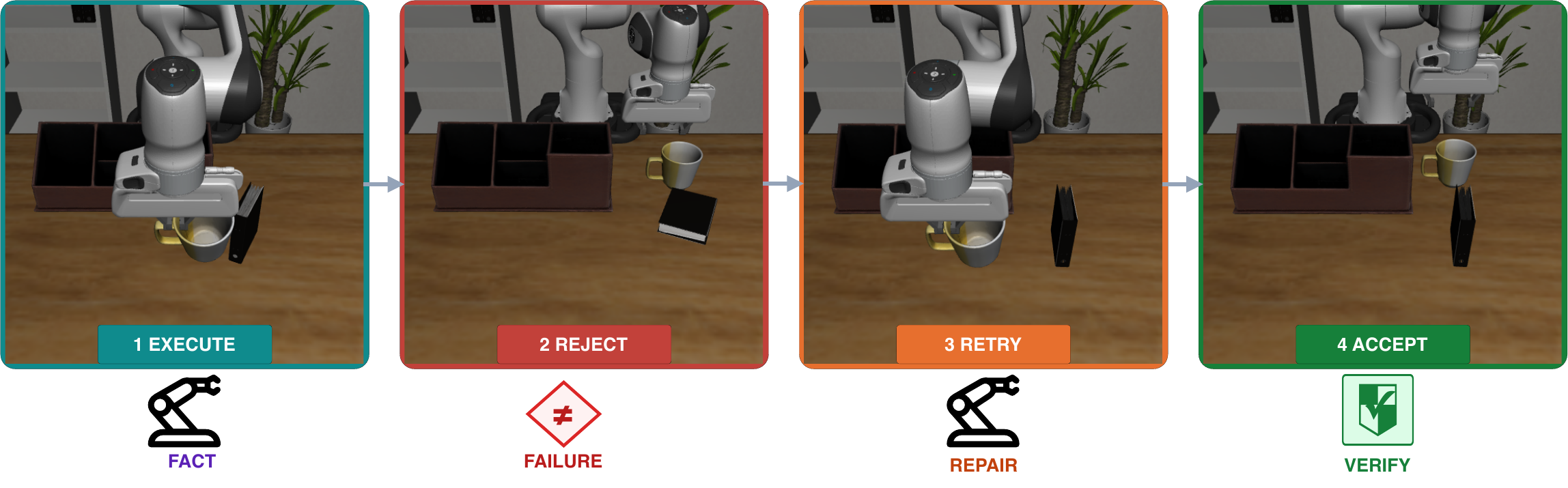}
\caption{Verifier-gated action repair on LIBERO-90 Task 76. The upper sequence executes candidate 1, rejects it, retries with candidate 2, and accepts the verified outcome. The icon rail summarizes the corresponding typed control path: an unsatisfied required relation instantiates a typed failure, the failure invokes an action-level repair, and the native task verifier determines acceptance.}
\label{fig:ai_harness_loop}
\end{figure}

Four records expose the interface state: facts store typed arguments and optional margins; routes retain predicate arguments and a correction-family label; applications distinguish simulator mutation from controller execution; and verification stores its query and outcome. The experiments do not measure audit quality or component interchangeability.

The quantitative LIBERO-Goal study uses one fixed learned source/joint \texttt{qpos}-delta mechanism with direct simulator-state application. In the separate Task 76 trace, an OSC-Pose pick-and-place controller and finite offset grid supply another candidate after rejection. This trace is qualitative, is not the quantitative \texttt{qpos} protocol, and does not establish a learned general-purpose recovery policy.

\subsection{Operational Robot Task Ontology}

We denote the task-facing interface by
\[
\mathcal I_g^{t,\sigma}=(\mathcal T,\mathcal A_g^{t,\sigma},\Pi_D,\Pi_R),\qquad
\mathcal T=(\mathcal C,\mathcal R,\Sigma_{\mathrm{type}}),\qquad
\mathcal A_g^{t,\sigma}=\phi_{\mathcal T}^{\sigma}(g,s_t^{\sigma}),
\quad \sigma\in\{\mathrm{pred},\mathrm{sim}\}.
\]
Here, $(\mathcal T,\mathcal A)$ is the ontological record layer, while $\Pi_D$ and $\Pi_R$ are external deterministic mappings. The class set $\mathcal C$ contains terms such as \emph{Object}, \emph{Region/Site}, \emph{Articulated Joint}, \emph{Relation}, \emph{Constraint}, \emph{Event/Margin}, \emph{Failure}, and \emph{Correction Route}. The relation vocabulary $\mathcal R$ contains typed predicates such as \texttt{right\_of(Object, Landmark)}, \texttt{inside(Object, Container)}, \texttt{contact(Object, Site)}, and \texttt{joint\_state(Joint, Threshold)}. The signature set $\Sigma_{\mathrm{type}}$ specifies declared arguments and type constraints. The source-specific grounding function $\phi_{\mathcal T}^{\sigma}$ compiles task specification $g$ and a provenance-tagged predicted or simulator state $s_t^{\sigma}$ into $\mathcal A_g^{t,\sigma}$. We omit $\sigma$ only where the source is already explicit. Diagnosis rules $\Pi_D$ and the route policy $\Pi_R$ are outside the TBox.

The TBox/ABox distinction follows standard description-logic terminology \citep{baader2007dlhandbook}, but is used here as a task-local interface: the TBox $\mathcal T$ is the reusable robot-task vocabulary, and the ABox is the current task instance. Required assertions are instantiated from BDDL-style task specifications \citep{liu2024libero}, while observed assertions and margins are grounded from simulator state or world-model output. Failure and route identifiers may use vocabulary declared by $\mathcal T$, but the assertion that a particular failure selects a particular route is produced by $\Pi_R$, not by the TBox. The current implementation directly evaluates enumerated schema keys, task bindings, and task-specific mappings. Thus, the TBox/ABox terminology describes the organization of these records; it does not imply general domain--range inference, open-world reasoning, or use of an external OWL reasoner.

The following expressions specify intended record-validity requirements for the enumerated task bindings; they are not theorems proved by the method or properties validated over arbitrary ABoxes. Predicate consistency requires declared argument types,
\[
r(x,y)\in\mathcal A\ \Rightarrow\
x{:}\operatorname{dom}_{\mathcal T}(r)\ \land\
y{:}\operatorname{range}_{\mathcal T}(r).
\]
Argument preservation requires a diagnosis to retain the violated predicate and its bindings,
\[
\mathrm{Required}_g(r(x,y))\land
\neg\mathrm{Sat}_{\mathcal T}(r(x,y),\mathcal A_g^t)
\Rightarrow
\mathrm{Failure}_t(\texttt{relation\_missing},r,x,y,m).
\]
Finally, compatible routing requires the external policy to preserve that record when it dispatches a mechanism,
\[
\Pi_R(\mathrm{Failure}(c,r,x,y,m))=\rho
\Rightarrow
\mathrm{Admits}(c,\rho)\land\mathrm{Supports}(\rho,r)\land
\mathrm{DeclaredTypeMatch}_{\mathcal T}(x,y,r).
\]
These requirements describe the intended structure of the enumerated bindings and mappings; the paper does not claim that a general runtime validator enforces them. The last implication concerns declared route compatibility only; it does not predict correction success.

The comparison is task-scoped: only assertions declared by $g$ enter the required set. Each diagnostic record retains the predicate identifier, typed argument bindings, required polarity, observed truth value, and, when available, a continuous margin. Facts unrelated to $g$ are therefore not treated as failures merely because they are absent from the current ABox. The route label and verifier outcome are stored as separate fields. The reported protocol counts a correction as successful only when the native task predicate is positive; a route label is not itself counted as success.

Table~\ref{tab:fact_schema} summarizes the declared vocabulary referenced by these record-format requirements. The route names in the table are external dispatch labels, not TBox axioms.

\begin{table}[!htbp]
\centering
\footnotesize
\renewcommand{\arraystretch}{1.05}
\caption{Core vocabulary of the operational robot task ontology.}
\label{tab:fact_schema}
\begin{tabularx}{\linewidth}{L{0.18\linewidth}L{0.28\linewidth}L{0.19\linewidth}Y}
\toprule
Class & Examples or signatures & Ontological role & Repair-loop use \\
\midrule
Object/Entity & bowl, plate, drawer, mug & predicate argument & bind task entities \\
Region/Site & stove\_front, caddy\_right, rack\_site & spatial landmark & ground target areas \\
Relation & on, inside, in\_region, right\_of & typed predicate & encode goal assertions \\
Action/Event & move, place, open, completed & transition/progress & describe proposed change \\
Constraint/Margin & contact, joint\_limit, target\_margin & condition/scalar & represent feasibility and threshold \\
Failure & predicate\_mismatch, relation\_missing, rack\_contact & diagnostic class & name the unmet condition \\
Correction Route & source\_joint\_pose, object\_relation\_predicate, rack\_contact\_site & repair family & select proposer and executor \\
\bottomrule
\end{tabularx}
\end{table}

\begin{figure}[H]
\centering
\includegraphics[width=\linewidth]{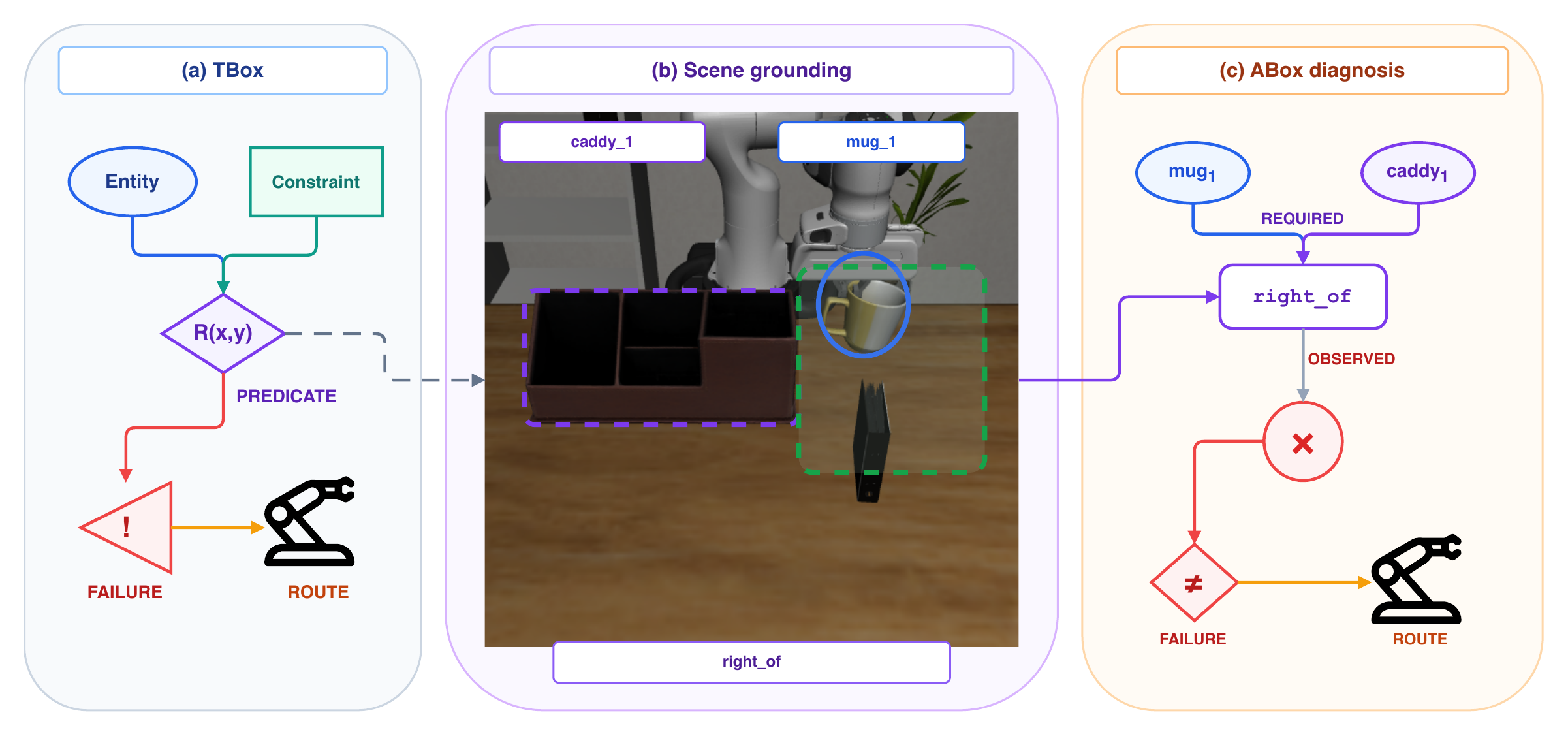}
\caption{Operational robot task ontology grounded in LIBERO-90 Task 76. \textbf{a}, The TBox supplies a compact grammar from typed entities and constraints to predicates, failures, and compatible routes. \textbf{b}, Scene grounding binds the mug, caddy, and target region. \textbf{c}, The task ABox requires \texttt{right\_of}; an absent observation instantiates a typed mismatch and selects an action route. Formal signatures and instance names are given in the text and Table~\ref{tab:fact_schema}; execution is shown in Figure~\ref{fig:ai_harness_loop}.}
\label{fig:operational_ontology}
\end{figure}

Figure~\ref{fig:operational_ontology} uses symbols for the control structure and leaves the full record names here. In panel a, only the entity types, predicate signatures, and constraints belong to the TBox; the failure and route glyphs show downstream interface context. For LIBERO-90 Task 76, the ABox binds \texttt{mug\_1:Object}, \texttt{caddy\_1:Landmark}, and the required assertion \texttt{right\_of(mug\_1,caddy\_1)}. A false observation yields the argument-preserving record
\[
\mathrm{relation\_missing}
(\mathrm{right\_of},\mathrm{mug}_1,\mathrm{caddy}_1),
\]
and the task-specific external mapping assigns \texttt{object\_relation\_predicate} while retaining both bindings. In the selected Task 76 example, an OSC-Pose controller executes the corresponding placement candidate. The native verifier determines whether that candidate is accepted; rejection leads to another candidate within the same illustrated route.

\subsection{Ontology Instantiation and Typed Diagnosis}

Diagnosis compares predicted or simulator-grounded facts with required facts. In PointMaze, relevant failures include excessive target distance and missing entity-on-target facts. In LIBERO-Goal (10 tasks), schema families include source/joint pose, object relation, rack contact, and residual verification. Relative to a scalar rank, the typed record adds the missing-condition name and argument bindings.

The schema associates each enumerated fact key with a task role. Task bindings identify the manipulated and target entities, while relation, constraint, optional margin, and failure fields describe the represented condition and unmet requirement.

Diagnosis is a comparison operation rather than a free-form explanation. For a required relation such as \texttt{in\_region(plate, stove\_front)}, the system records whether the relation is present and whether the corresponding margin meets the task threshold. Missing relations and violated margins are mapped to named failure categories. The resulting record can index an enumerated route mapping; the quantitative LIBERO-Goal experiment, however, fixes the source/joint route and does not compare this mapping with scalar-only selection.

The schema distinguishes four task-state diagnosis families: predicate or relation mismatch, continuous margin or state mismatch, contact mismatch, and articulation or joint-state mismatch. An execution failure is different: it is the native verifier's negative outcome after a proposed application. This taxonomy explains the semantics of existing fact records and qualitative traces; no category-frequency claim is made from it.

\subsection{State-Correction Router}

\begin{table}[!htbp]
\centering
\footnotesize
\renewcommand{\arraystretch}{1.05}
\caption{Typed gates and state-correction families. Only the learned source/joint \texttt{qpos}-delta route contributes to the quantitative LIBERO Goal10 state-correction result; the other rows define schema and qualitative routing semantics.}
\label{tab:router}
\begin{tabularx}{\linewidth}{L{0.23\linewidth}L{0.22\linewidth}L{0.25\linewidth}Y}
\toprule
Typed gate & State-correction family & Example tasks & Intended effect \\
\midrule
source\_joint\_pose & learned source/joint delta & drawers, stove, placement & correct object or joint state \\
object\_relation\_predicate & object-relative correction & bowl on plate, containment & satisfy target relation \\
rack\_contact\_site & rack-local pose correction & bottle on rack & satisfy contact predicate \\
source\_joint\_residual & verifier-checked candidate & source/joint residuals & retain predicate-valid state \\
\bottomrule
\end{tabularx}
\end{table}

Table~\ref{tab:router} summarizes the typed dispatch layer. In the schema, source/joint failures are associated with corrections parameterized in source or joint coordinates, object-relation failures with object-relative corrections, and rack-contact failures with a rack-local frame. These associations define route labels and coordinate families; they are not results from online selection among learned heads. Only the learned source/joint \texttt{qpos}-delta route contributes to the quantitative multi-seed result.

A route label that matches the declared schema does not establish that its proposal will satisfy the target predicate. The diagnostic record and verifier outcome are stored separately, and only the native predicate determines acceptance.

\subsection{Failure-Gated State-Correction Protocol}

The quantitative LIBERO-Goal protocol replays a 25-step demonstration window and queries the native success predicate. Only failed windows enter the correction gate. For this experiment, one fixed learned head consumes the recorded window features $x_j$ and proposes a \texttt{qpos} delta:
\[
q_j=P_{\mathrm{fixed}}(x_j),\qquad
s'_j=\mathrm{Apply}(s_j,q_j),\qquad
v_j=\mathbb I[\texttt{check\_success}(s'_j)=1].
\]
Here, $x_j$ contains the task and state features supplied to the fixed proposer, and $\mathrm{Apply}$ directly writes simulator \texttt{qpos}. The typed diagnostic record annotates the source/joint failure family, but the quantitative experiment does not dynamically choose among multiple route-specific heads. Direct \texttt{qpos} application is not a sequence of robot actions and does not test whether a policy can realize the transition under environment dynamics. The failure gate leaves every successful replay unchanged and evaluates corrections only on replay failures. Table~\ref{tab:protocol} fixes the interpretation.

\begin{table}[!htbp]
\centering
\footnotesize
\renewcommand{\arraystretch}{1.05}
\caption{LIBERO Goal10 evaluation protocol. Sampling seeds change the evaluated window sample while the learned checkpoint remains fixed.}
\label{tab:protocol}
\begin{tabularx}{0.94\linewidth}{L{0.34\linewidth}Y}
\toprule
Field & Setting \\
\midrule
Evaluation unit / horizon & 25-step demonstration window, $H=25$ \\
Windows & 50 per task, 500 per sampling seed; seeds 0--3 \\
Checkpoint & One fixed learned source/joint \texttt{qpos}-delta head \\
Proposal source & Learned delta on replay failures only \\
Verifier & Native LIBERO \texttt{check\_success} \\
State application & Direct simulator \texttt{qpos} mutation \\
Independent training seeds & Not evaluated \\
Held-out demonstration split & Not used \\
Full-episode policy evaluation & Not evaluated \\
\bottomrule
\end{tabularx}
\end{table}

The reported corrected-window aggregate is constructed as follows. Replay-success windows retain their original outcome. For each replay failure, the fixed checkpoint proposes one source/joint correction and the protocol queries the native predicate after direct state application. Corrected success equals the retained replay successes plus replay-failure windows that satisfy the predicate after this application.

This construction evaluates the checkpoint on replay-failure windows and carries replay-success outcomes forward by definition. It remains a window-level diagnostic protocol because the correction is not rolled out as an action sequence and does not restart a complete task episode.

\section{Experiments}

We report an aligned PointMaze comparison, a sampled-window LIBERO-Goal evaluation of one fixed simulator-state correction checkpoint, selected qualitative route traces, and fixed-registry LIBERO-Plus results. These protocols are interpreted separately; the qualitative traces are not correction trials.

\subsection{PointMaze}

The PointMaze record contains distance to the goal, target margin, progress, success, state validity, and failure type.

The aligned comparison uses the same random-state setting and 50-trial evaluation for DINO-WM, EV-WM, and Onto-EV-WM. Success rate measures whether the final state satisfies the task, whereas mean state distance retains proximity information for both successful and unsuccessful trials. The final larger-budget row is reported separately because it changes the candidate budget and prioritizes success during selection.

\begin{table}[!htbp]
\centering
\footnotesize
\renewcommand{\arraystretch}{1.05}
\caption{PointMaze random-state planning results, including the published EV-WM baseline \citep{wang2026evwm}. The first three rows use the same 50-trial setting. The final row changes both search budget and success-first selection and is not directly aligned.}
\label{tab:pointmaze}
\begin{tabularx}{0.92\linewidth}{lYcc}
\toprule
Method & Setting & Success rate & Mean state distance \\
\midrule
DINO-WM baseline & aligned & 0.90 & 0.93568 \\
EV-WM & aligned & 0.94 & 0.90573 \\
Onto-EV-WM & aligned & 0.94 & \best{0.61177} \\
Onto-EV-WM + diagnosis-guided search & larger budget, success-first & \best{1.00} & 0.86504 \\
\bottomrule
\end{tabularx}
\end{table}

Under the aligned setting, EV-WM and Onto-EV-WM both report a success rate of 0.94. Their mean final-state distances are 0.90573 and 0.61177, respectively, corresponding to a 32.5\% lower observed value for Onto-EV-WM in these 50 trials. The two observed diagnostic categories are represented explicitly in the task record: target-distance failure denotes a violated continuous goal margin, while missing-entity-on-target denotes an unsatisfied discrete spatial assertion. Under the separately budgeted success-first configuration, 50/50 trials satisfy the task predicate and the mean final-state distance is 0.86504. Because the search budget and selection rule differ, neither quantity in this row is interpreted as a matched effect relative to the aligned rows.

The aligned 50-trial sample does not establish statistical reliability, and the separately budgeted row does not establish budget-matched superiority.

\subsection{LIBERO-Goal (10 Tasks)}

LIBERO-Goal contains 10 manipulation task specifications spanning articulation, placement, spatial regions, support relations, rack contact, and stove activation \citep{liu2024libero}. For each evaluation-sampling seed, we sample fifty 25-step demonstration windows per task (500 in total). Demonstration replay provides the pre-correction reference rather than a separate baseline method. The corrected-window metric retains replay-success outcomes and queries \texttt{check\_success} after direct state application on replay-failure windows.

\paragraph{Ontology contribution in LIBERO-Goal.}
The ontology compiles each structured goal into typed required assertions, grounds the sampled simulator state into an observed ABox, and records an unsatisfied predicate with its arguments, polarity, available margin, declared fixed route, and separate verifier outcome. The learned head supplies the \texttt{qpos} proposal; the ontology supplies the diagnosis and correction record that connects the failed condition to predicate-gated acceptance. The reported rate therefore measures the complete ontology-grounded correction configuration, not the learned head alone.

The evaluation unit is a sampled window, not a full episode. Each seed changes the sampled windows while preserving the checkpoint and protocol, so the four seeds describe evaluation-sample variation rather than independent training. The overlapping label-construction and evaluation pools do not test unseen-demonstration generalization. Across the four samples, replay success is $\LiberoBeforeMeanPct\!\pm\!\LiberoBeforeStdPct\%$ and corrected success is $\LiberoAfterMeanPct\!\pm\!\LiberoAfterStdPct\%$; pooled over \LiberoTotalWindows{} windows, \LiberoRescuedCount{} of \LiberoBaselineFailureCount{} replay failures satisfy \texttt{check\_success} after direct simulator-state application.

Replay success ranges from 428/500 to 440/500, corrected success from 469/500 to 472/500, and recovered failures from 32 to 43 per sample. These descriptive ranges are not confidence intervals over independently trained systems. Table~\ref{tab:libero_staged} gives the per-seed and pooled counts.

\begin{table}[!htbp]
\centering
\footnotesize
\caption{LIBERO Goal10 sampled-window results across four evaluation-sampling seeds. Event reports the event-score CEM result; ``Rescued'' counts replay failures converted to predicate-valid outcomes by direct simulator-state correction. Mean and sample standard deviation are calculated across evaluation-sampling seeds, not independently trained models.}
\label{tab:libero_staged}
\begin{tabular}{cC{0.16\linewidth}C{0.17\linewidth}C{0.17\linewidth}C{0.22\linewidth}}
\toprule
Sampling seed & Event success & Replay success & Corrected success & Rescued / replay failures \\
\midrule
0 & 392/500 & \LiberoSeedZeroBefore/500 & \LiberoSeedZeroAfter/500 & \LiberoSeedZeroRescued/\LiberoSeedZeroFailures \\
1 & 379/500 & \LiberoSeedOneBefore/500 & \LiberoSeedOneAfter/500 & \LiberoSeedOneRescued/\LiberoSeedOneFailures \\
2 & 398/500 & \LiberoSeedTwoBefore/500 & \LiberoSeedTwoAfter/500 & \LiberoSeedTwoRescued/\LiberoSeedTwoFailures \\
3 & 401/500 & \LiberoSeedThreeBefore/500 & \LiberoSeedThreeAfter/500 & \LiberoSeedThreeRescued/\LiberoSeedThreeFailures \\
\midrule
Mean $\pm$ s.d. & $78.50\!\pm\!1.95\%$ & $\LiberoBeforeMeanPct\!\pm\!\LiberoBeforeStdPct\%$ & $\LiberoAfterMeanPct\!\pm\!\LiberoAfterStdPct\%$ & --- \\
Pooled & 1570/2000 & \LiberoPooledBeforeCount/\LiberoTotalWindows & \LiberoPooledAfterCount/\LiberoTotalWindows & \LiberoRescuedCount/\LiberoBaselineFailureCount{} (\LiberoRecoveryPct\%) \\
\bottomrule
\end{tabular}
\end{table}

\begin{figure}[htbp]
\centering
\includegraphics[width=\linewidth]{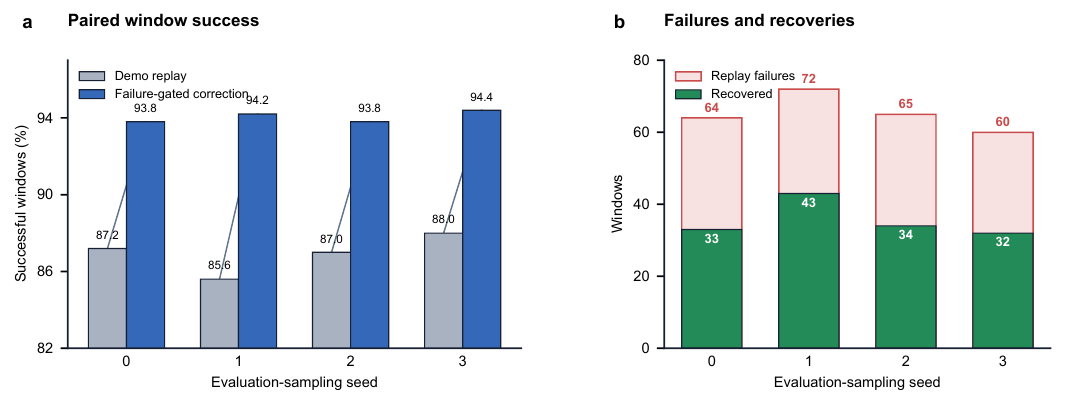}
\caption{LIBERO Goal10 sampled-window evaluation. \textbf{a}, Paired replay and failure-gated corrected success for four evaluation-sampling seeds. \textbf{b}, Replay failures and the subset recovered by the same fixed learned correction checkpoint. Seeds vary the sampled evaluation windows rather than model training; the protocol uses neither held-out demonstrations nor full episodes.}
\label{fig:libero_staged_gain}
\end{figure}

Figure~\ref{fig:libero_staged_gain} reports paired replay/corrected success and the replay-failure denominators. The corrected rate is higher for every sample, but the typed schema and learned correction head are not isolated in this protocol. The values therefore characterize the complete ontology-grounded correction configuration rather than an ontology-only numerical effect.

\paragraph{Qualitative diagnosis traces.}

The traces illustrate the distinction between diagnosis and verification. A gate names a correction family but does not certify completion; only the task predicate determines the reported outcome.

\begin{figure}[htbp]
\centering
\includegraphics[width=0.58\linewidth]{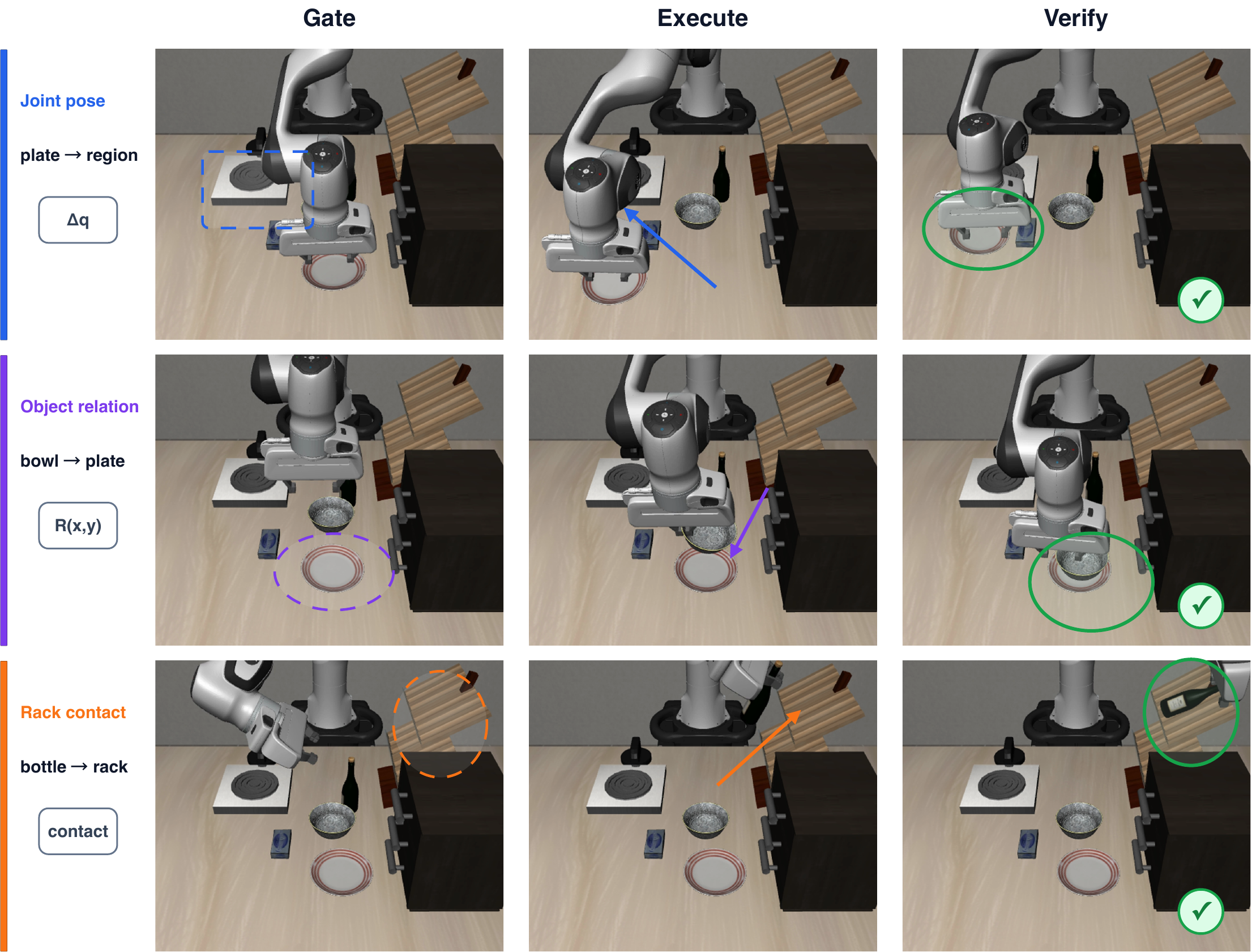}
\caption{Three LIBERO Goal10 evidence chains. The row label and compact route glyph identify the admitted correction family; dashed outlines denote required regions, colored arrows illustrate execution direction, and green contours with check marks denote predicate-supporting evidence. The selected full-demonstration traces explain schema semantics; they are not additional learned state-correction trials.}
\label{fig:robot_trace}
\end{figure}

Each row in Figure~\ref{fig:robot_trace} follows the same gate--execute--verify reading order. The compact left-hand glyph names the route without repeating its implementation record; the gate image marks the required predicate or region, the execution image illustrates the coordinate family associated with the diagnosis, and the final image marks predicate-supporting evidence. In full schema terms, the three chains are missing \texttt{in\_region} $\rightarrow$ \texttt{source\_joint\_pose}, missing \texttt{on} $\rightarrow$ \texttt{object\_relation\_predicate}, and missing \texttt{contact} $\rightarrow$ \texttt{rack\_contact\_site}. These labels expand the compact glyphs without altering the qualitative evidence shown in the figure.

These traces are explanatory rather than quantitative. They are selected full-demonstration frames and do not add learned-correction trials to Table~\ref{tab:libero_staged}. Their role is to show how the same schema represents spatial regions, object relations, and local contact conditions without conflating a route label with a successful action sequence.

\subsection{LIBERO-Plus}

LIBERO-Plus expands the four standard LIBERO suites into a fixed registry of 10,030 perturbation tasks \citep{fei2026liberoplus}. The task set used here comprises 2,519 LIBERO-10 (\texttt{libero\_10}) tasks, 2,591 LIBERO-Goal (\texttt{libero\_goal}) tasks, 2,518 LIBERO-Object (\texttt{libero\_object}) tasks, and 2,402 LIBERO-Spatial (\texttt{libero\_spatial}) tasks. The suite definitions emphasize longer multi-stage tasks, goal variation, object variation, and spatial variation, respectively.

The reporting unit is a registered task, and the four suite totals define the fixed denominator. The aggregate Onto-EV-WM success rate is 8,526/10,030 (85.00\%). The suite-level success rates are above 91\% on LIBERO-Goal, LIBERO-Object, and LIBERO-Spatial, and 65.98\% on LIBERO-10. The internal DINO-WM+CEM baseline \citep{rubinstein1999cem} disables ontology scoring and completes one seed-0 online trial for every registered task. Its comparison with the full Onto-EV-WM configuration is system-level and does not by itself isolate the causal contribution of ontology scoring.

Table~\ref{tab:libero_plus_external} places the result alongside selected VLAs, world-action models, and geometry-enhanced policies. The literature values are reproduced from the LIBERO-Plus suite table reported by GAM, which combines re-evaluated available checkpoints with values imported from the corresponding published benchmarks \citep{han2026gam}. The rows share the reported suite breakdown but do not constitute controlled head-to-head runs with identical checkpoints, seeds, or execution protocols.

\begin{table}[!htbp]
\centering
\footnotesize
\setlength{\tabcolsep}{4.5pt}
\renewcommand{\arraystretch}{1.02}
\caption{LIBERO-Plus comparison with representative methods. Average is the reported aggregate benchmark success rate; because suite sizes differ, it is not the unweighted mean of the four suite percentages. The best and second-best results are shown in bold and underlined, respectively.}
\label{tab:libero_plus_external}
\begin{tabular}{lccccc}
\toprule
Model & Long & Goal & Object & Spatial & Average \\
\midrule
$\pi_{0.5}$ & 77.9 & 79.3 & 89.9 & \underline{92.0} & 84.6 \\
OpenVLA-OFT & 66.4 & 63.0 & 66.5 & 84.0 & 69.6 \\
$\pi_{0}$ & 61.3 & 63.4 & 74.7 & 78.6 & 69.3 \\
RIPT-VLA & 67.5 & 58.0 & 64.3 & 85.8 & 68.4 \\
$\pi_{0}$-FAST & 43.4 & 57.5 & 72.7 & 74.4 & 61.6 \\
UniVLA & 39.9 & 40.7 & 36.7 & 55.5 & 42.9 \\
NORA & 36.3 & 38.8 & 34.4 & 47.6 & 39.0 \\
OpenVLA & 14.3 & 15.1 & 14.0 & 19.4 & 15.6 \\
\addlinespace[3pt]
Cosmos-Policy & \textbf{81.0} & 73.5 & 88.3 & 87.3 & 82.4 \\
Fast-WAM & 41.1 & 39.2 & 71.2 & 54.4 & 50.0 \\
WorldVLA & 8.2 & 31.8 & 28.6 & 32.5 & 25.0 \\
\addlinespace[3pt]
GAM & \underline{78.0} & \underline{80.4} & \underline{90.6} & \textbf{93.4} & \textbf{85.5} \\
$\pi_{0.5}$ + ROCKET & 36.7 & 39.7 & 53.9 & 31.5 & 47.5 \\
\midrule
DINO-WM + CEM (no ontology) & 30.73 & 64.22 & 74.15 & 77.85 & 61.57 \\
\textbf{Onto-EV-WM (ours)} & 65.98 & \textbf{91.39} & \textbf{91.38} & 91.38 & \underline{85.00} \\
\bottomrule
\end{tabular}
\end{table}

Of the 1,504 unsuccessful registered tasks, 857 (56.98\%) belong to LIBERO-10, compared with 223 in LIBERO-Goal, 217 in LIBERO-Object, and 207 in LIBERO-Spatial. Thus, although LIBERO-10 accounts for roughly one quarter of the registry, it contains more than half of the unsuccessful tasks. The count distribution identifies only the suite with the largest residual count; without trace-level labels, it supports no finer causal attribution.

\section{Discussion}

\subsection{Interface Roles and Observed System Results}

The interface stores typed predicates, source-tagged ABoxes, diagnostic fields, route records, and verifier outcomes; a route label is not counted as success. The experiments evaluate complete Onto-EV-WM configurations rather than an ontology-only component.

The protocol-specific observations are as follows: the aligned PointMaze rows both report 0.94 success; the LIBERO-Goal ontology-grounded correction configuration represents each failed predicate, its arguments, fixed route, and verifier outcome around one checkpoint and reports 93.8\% on seed 0 and $\LiberoAfterMeanPct\!\pm\!\LiberoAfterStdPct\%$ across window samples; and the fixed LIBERO-Plus registry reports 85.00\%, with 65.98\% on LIBERO-10 and rates above 91\% on the other suites. These values describe the complete configurations and are not combined into one causal estimate.

Compared with a scalar verifier output, the typed record additionally stores the missing predicate, arguments, margin, correction family, application mode, and acceptance decision. Audit quality and component-level causal shares are not measured.

\subsection{Limitations and Scope}

The aligned PointMaze comparison contains 50 trials, and its 100\% row changes both search budget and selection. LIBERO-Goal uses sampled windows from a pool overlapping label construction, one fixed checkpoint, and direct simulator-state changes; its four seeds are not independent training runs, and the protocol is not full-episode policy evaluation. LIBERO-Plus is a cross-architecture fixed-registry comparison, not a component-isolation study. All evidence is simulation-based. Held-out generalization, component-level causality, and real-robot recovery remain outside the claim.

\section{Conclusion}

Onto-EV-WM adds typed grounding, diagnosis, route records, and predicate-gated acceptance to EV-WM. In aligned PointMaze, EV-WM and Onto-EV-WM both report 0.94 success, with mean final-state distances of 0.90573 and 0.61177; the separate larger-budget run reaches 50/50 success (100\%). In LIBERO-Goal, the ontology represents the failed condition and declared correction contract around one fixed \texttt{qpos}-delta checkpoint; this complete configuration reports 469/500 (93.8\%) on seed 0 and $\LiberoAfterMeanPct\!\pm\!\LiberoAfterStdPct\%$ across four evaluation-sampling seeds. On the fixed LIBERO-Plus registry, the Onto-EV-WM configuration reports 8,526/10,030 (85.00\%).

These are integrated-system results under simulation protocols. The quantitative correction applies simulator state directly, while controller-mediated execution appears only in selected qualitative traces. The results do not establish held-out generalization, an ontology-only causal effect, real-robot transfer, or an end-to-end learned correction policy.

\bibliographystyle{abbrvnat}
\begingroup
\setlength{\parskip}{0pt}
\interlinepenalty=10000
\raggedright
\bibliography{references}
\endgroup

\end{document}